\documentclass[sigconf]{acmart}

\usepackage{amsmath}
\usepackage{amssymb}
\usepackage{bm}
\usepackage{booktabs}
\usepackage{adjustbox}
\usepackage{multirow}
\usepackage[inline]{enumitem}
\usepackage{graphicx} 
 
\usepackage{tikz}
\usetikzlibrary{positioning, fit, mindmap, trees, calc,tikzmark,shapes}
\usetikzlibrary{shapes.arrows, fadings, automata,tikzmark,decorations.pathreplacing,patterns}
\usetikzlibrary{shapes.multipart}
\usetikzlibrary{arrows.meta}
\usetikzlibrary{intersections}
\usetikzlibrary{fit}

\usepackage{pgfplots}
\pgfplotsset{compat=1.14} 
\usetikzlibrary{pgfplots.groupplots}

\usepackage{xcolor}
\definecolor{mygray}{gray}{.9}
\definecolor{size}{RGB}{169, 198, 207}
\definecolor{casual}{RGB}{221,237,194}
\definecolor{female}{RGB}{255,212,181}
\definecolor{hot}{RGB}{255,171,166}

\definecolor{bgc}{RGB}{245,245,245}

\definecolor{persian_green}{RGB}{0, 166, 141}
\definecolor{macaroni_and_cheese}{RGB}{255, 196, 131}
\definecolor{bittersweet}{RGB}{255, 112, 98}

\definecolor{line1}{RGB}{226, 58, 89}
\definecolor{line2}{RGB}{68 ,173,173}
\definecolor{line3}{RGB}{60, 80, 107}
\definecolor{line5}{RGB}{252,200,43}
\definecolor{line4}{RGB}{15, 94, 140}
\definecolor{line6}{RGB}{173,107,53}

\definecolor{tuatara}{RGB}{67, 67, 67}
\definecolor{aluminum}{RGB}{153,153,153}
\definecolor{silver}{RGB}{191,191,191}
\definecolor{platinum}{RGB}{228,228,228}
\definecolor{mercury}{RGB}{230,230,230}
\definecolor{gallery}{RGB}{240,240,240}
\definecolor{free_speech_aquamarine}{RGB}{0, 156, 114}
\definecolor{sun_shade}{RGB}{255, 144, 68}
\definecolor{fern}{RGB}{101,197,117}
\definecolor{french_blue}{RGB}{0, 112, 182}
\definecolor{matisse}{RGB}{25, 104, 167}
\definecolor{sushi}{RGB}{117, 168, 47}
\definecolor{shakespeare}{RGB}{85, 154, 193}
\definecolor{egg_shell}{RGB}{238, 234, 215}
\definecolor{carnation}{RGB}{245, 80, 86}
\definecolor{flamingo}{RGB}{237, 88, 85}
\definecolor{jet_stream}{RGB}{188, 214, 210}
\definecolor{jelly_bean}{RGB}{45, 126, 150}
\definecolor{tree_poppy}{RGB}{246, 154, 27}
\definecolor{deep_carmine_pink}{RGB}{236, 50, 67}
\definecolor{copper_rust}{RGB}{155, 64, 74}
\definecolor{midnight}{RGB}{0, 29, 50}
\definecolor{chilean_fire}{RGB}{215, 87, 44}
\definecolor{puerto_rico}{RGB}{94, 194, 166}
\definecolor{japanese_laurel}{RGB}{53, 116, 40}
\definecolor{fire_engine_red}{RGB}{206, 37, 51}
\definecolor{ku_crimson}{RGB}{243, 0, 25}
\definecolor{turmeric}{RGB}{211, 178, 76}
\definecolor{tahiti_gold}{RGB}{223, 102, 36}
\definecolor{outrageous_orange}{RGB}{255, 100, 45}
\definecolor{crusta}{RGB}{254, 127, 44}
\definecolor{safety_orange}{RGB}{254, 106, 0}
\definecolor{pigment_green}{RGB}{0, 175, 79}
\definecolor{jaffa}{RGB}{240, 131, 58}
\definecolor{jet_stream}{rgb}{0.69,0.61,0.85}
\definecolor{jelly_bean}{rgb}{0.47,0.32,0.66}
\definecolor{azalea}{RGB}{251, 196, 196}
\definecolor{sundown}{RGB}{249, 180, 181}
\definecolor{light_coral}{RGB}{244, 127, 123}
\definecolor{wewak}{RGB}{244, 143, 150}

\definecolor{biscay}{RGB}{44, 62, 80}
\definecolor{carmine_pink}{RGB}{231, 76, 60}
\definecolor{athens_gray}{RGB}{236, 240, 241}
\definecolor{celestial_blue}{RGB}{52, 152, 219}
\definecolor{curious_blue}{RGB}{41, 128, 185}
\definecolor{my_sin}{RGB}{255, 176, 59}
\definecolor{viridian}{RGB}{70, 137, 102}
\definecolor{tomato}{RGB}{255, 97, 56}
\definecolor{mountain_meadow}{RGB}{0, 163, 136}
\definecolor{padua}{RGB}{121, 189, 143}
\definecolor{killarney}{RGB}{56, 113, 66}
\definecolor{ocean_green}{RGB}{79, 176, 112}
\definecolor{pastel_green}{RGB}{107, 227, 135}
\definecolor{chinook}{RGB}{163, 232, 178}
\definecolor{cosmic_latte}{RGB}{222, 247, 229}
\definecolor{chateau_green}{RGB}{69, 191, 85}
\definecolor{RoyalBlue}{RGB}{69, 191, 85}

\usepackage{sistyle}
\SIthousandsep{,}

\setcopyright{rightsretained}

\acmDOI{10.475/123_4}

\acmISBN{123-4567-24-567/08/06}

\acmConference[WOODSTOCK'97]{ACM Woodstock conference}{July 1997}{El
  Paso, Texas USA}
\acmYear{1997}
\copyrightyear{2016}

\acmArticle{4}
\acmPrice{15.00}

\begin{document}
\title{Compositional Network Embedding}

\author{Tianshu Lyu}
\affiliation{%
  \institution{Peking University}
  \department{Department of Machine Intelligence}
}
\email{lyutianshu@pku.edu.cn}

\author{Fei Sun}
\affiliation{Alibaba Group} 
\email{ofey.sf@alibaba-inc.com}

\author{Peng Jiang}
\affiliation{Alibaba Group}
\email{jiangpeng.jp@alibaba-inc.com}

\author{Wenwu Ou}
\affiliation{Alibaba Group}
\email{santong.oww@taobao.com}

\author{Yan Zhang}
\affiliation{%
  \institution{Peking University}
  \department{Department of Machine Intelligence}
}
\email{zhy@cis.pku.edu.cn}

\begin{abstract}
Network embedding has proved extremely useful in a variety of network analysis tasks such as node classification, link prediction, and network visualization.
Almost all the existing network embedding methods learn to map the node IDs to their corresponding node embeddings.
This design principle, however, hinders the existing methods from being applied in real cases.
Node ID is not generalizable and, thus, the existing methods have to pay great effort in cold-start problem.
The heterogeneous network usually requires extra work to encode node types, as node type is not able to be identified by node ID.
Node ID carries rare information, resulting in the criticism that the existing methods are not robust to noise.

To address this issue, we introduce \textit{Compositional Network Embedding}, a general \textit{inductive} network representation learning framework that generates node embeddings by combining node features based on the ``\textit{principle of compositionally}''.
Instead of directly optimizing an embedding lookup based on arbitrary node IDs, we learn a composition function that infers node embeddings by combining the corresponding node attribute embeddings through a graph-based loss. 
For evaluation, we conduct the experiments on link prediction under four different settings.
The results verified the effectiveness and generalization ability of compositional network embeddings, especially on unseen nodes.

\end{abstract}



\maketitle

\section{Introduction}
Recently, there has been a surge of work to represent nodes as low-dimensional dense vectors, called \textit{Network Embedding}.
Unlike traditional approaches relying on hand-crafted features to encode graph topology, network embedding automatically learns to encode graph structure into low-dimensional vectors using deep learning \cite{DBLP:conf/kdd/GroverL16,DBLP:conf/www/TangQWZYM15} or dimensionality reduction \cite{Belkin:2001:LES:2980539.2980616,Ahmed:2013:DLN:2488388.2488393}.

As the downstream machine learning tasks usually involve the predictions relevant to the nodes, the existing methods usually take node IDs as input and map them to a latent space.
However, we are afraid that many challenges the existing methods have to face are originated from this designing principle. 

First of all, node ID is not generalizable and, therefore, most of the existing methods cannot infer the embeddings of the unseen nodes that do not appear in the training phase.
Dynamic network embeddings generate unseen node embeddings in incremental \cite{ijcai2018-288} or inductive way \cite{DBLP:journals/corr/KipfW16,DBLP:conf/nips/HamiltonYL17}.
However, both of these two kinds of methods require the unseen nodes have connections to the observed network, which is impossible in some real-world problems (\textit{e.g.}, cold- start items in recommendation system).

The second challenge is particularly for the heterogeneous network.
Node IDs do not carry node types information inherently.
Consequently, it is inappropriate for the embedding methods learning the mapping function between node IDs and types.
A conventional way for heterogeneous network embedding is to project different types of nodes to different latent spaces.
This kind of heterogeneous network embeddings \cite{DBLP:conf/kdd/ChangHTQAH15,dong2017metapath2vec,qu2017attention} have to put great effort in aligning embeddings of different spaces.

The last one is the balance between network topology sensitivity and robustness.
It is ideal for the embeddings to preserve the structural information and be robust to tiny topology changes.
However, each edge is represented by a pair of node IDs, giving little hint about the edge itself.
Therefore, robustness is hardly discussed in the network embedding fields \cite{anonymous2019data}.
The methods based on random-walk sampling \cite{DBLP:conf/kdd/GroverL16,DBLP:conf/www/TangQWZYM15} are robust to edge existences but not because of the rationality.
On the contrary, the node embeddings relying on aggregating neighborhood attributes \cite{DBLP:journals/corr/KipfW16,DBLP:conf/nips/HamiltonYL17} might be seriously affected by the wrong edges \cite{pmlr-v80-dai18b,zugner2018adversarial}.

In this paper, we draw inspiration from the ``\textit{principle of compositionally}'' \cite{Frege:1982:Uber,battaglia2018relational,chomsky2014aspects,Mitchell:Composition}, an influential theory in NLP area, in order to inherently tackle the above problems. 
It states the meaning of a complex expression is determined by the meanings of its constituent expressions and the rules used to combine them.
For example, morphemes are combined into words, words into phrases, phrases into sentences, and sentences into paragraphs \cite{Mitchell:Composition,Li:Hierarchical:ACL2015,Kiros:Skip:NIPS2015}.
By analogy, we model the network in a compositional way, by deriving the node embeddings from their attributes.

Instead of learning a distinct embedding vector for each node ID, CNE trains a composition function that learns to deriving the node embedding by combining the corresponding node attribute embeddings which are shared across all nodes in the network.
Thus, at test time, we can generate embeddings for the unseen nodes by applying the learned composition function to their attributes.
Specifically, CNE learns the attribute embeddings and the composition function by considering the node proximity indicating from the graph, which is flexibly captured by random walks and sliding window.
The node embeddings, as the intermediate results of the framework, are encouraged to be more similar by the unsupervised graph-based loss function when the nodes are in close proximity.

Compared with classic network embedding methods, Compositional Network Embedding (CNE) is innately with three advantages:

\textbf{Able to infer the embeddings of the unseen node. }
Once CNE is well-trained, the embeddings of new nodes could be inferred with the node attributes as input.

\textbf{Easy to apply to heterogeneous network.} 
Different types of nodes correspond to different node attributes and composition methods.
The type differences are naturally captured by CNE.

\textbf{Robust to less informative edges.}
CNE models the network topology on the basis of sharing node attributes and composition methods, which act as powerful regularizations.
CNE is not sensitive to the edges that related to two nodes with rarely co-occurred attributes (\textit{e.g.} a Real Madrid fan clicked a FC Barcelona team jersey).

We evaluate our models on link prediction under different settings, including the prediction of missing edges, edges of unseen nodes, multi-edge-type edges, and multi-node-type edges. 
The experimental results demonstrate that CNE possesses higher expressive capability, stronger generalization capacity, and more flexibility on heterogeneous network.

\section{Related Work}
In this section, we briefly review the previous network embedding methods.
According to the design principles, we classify these methods into two categories, non-compositional and compostional methods.
Network embeddings methods are supposed to map each node to a low-dimensional vector by an \textit{encoder} \cite{Hamilton2017RepresentationLO}.
The \textit{encoder} of non-compositional methods take  indivisible data (\textit{i.e.} node ID) as input, while compositional methods requires an aggregation of information as input.

\subsection{Non-compositional Methods}
Unlike early dimensionality reduction methods usually preserving lower order proximity of networks which is often very sparse, recent network embedding methods attempt to model higher order proximity for effectiveness.
The recent successuful network embedding algorithms are inspired by the emergence of neural language models \cite{Bengio:2003:NPL:944919.944966} and word embedding methods like Skip-Gram \cite{Mikolov:2013:DRW:2999792.2999959}.
DeepWalk \cite{DBLP:conf/kdd/PerozziAS14} first bridges network embeddings and word embeddings by treating nodes as words and applying Skip-Gram to those generated short random walks to learn node embeddings.
Inspired by the success of DeepWalk, plenty of algorithms are proposed to learn node embeddings using random walk statistics \cite{Cao:2015:GLG:2806416.2806512,DBLP:conf/www/TangQWZYM15,DBLP:conf/kdd/GroverL16,Lyu:2017:ENE:3132847.3132900,Ou:2016:ATP:2939672.2939751,Perozzi:2017:DWS:3110025.3110086}.
Intuitively, these methods measure the graph proximity using different random walk strategies, and achieve superior performance in a number of settings \cite{DBLP:journals/kbs/GoyalF18}.

We treat these methods as non-compositional methods, as Skip-Gram model takes word ID as input, and on the network, they simply learn an embedding lookup based on node IDs.
Although the design principle is neat, non-compositional methods are inherently transductive.
They cannot naturally generate embeddings for unseen nodes unless additional rounds of optimization are performed on these nodes \cite{ijcai2018-288}.
Moreover, node attributes are failed to be leveraged, which severely limits the representation capacity.

\subsection{Compositional Methods}
It is a trend that the recent researches tackle the above problems in a compositional way,
although ``\textit{principle of compositionally}'' may not be explicitly claimed.
GraphSAGE \cite{DBLP:conf/nips/HamiltonYL17}, Graph Convolutional Networks (GCN) \cite{DBLP:journals/corr/KipfW16}, and DeepGL \cite{DBLP:conf/www/RossiZA18} generate the target node embedding by aggregating the neighbor feature embeddings within several hops.
Graph Attention Networks (GAT) \cite{DBLP:journals/corr/abs-1710-10903} stacks graph attention layers, making the node attend over its direct neighbors' features.
The recently proposed Graph Network (GN) formalism \cite{battaglia2018relational} unifies the above methods into one framework but with different configurations.
The first three methods are classified into Message-Passing Neural Networks family, making node features propagate on the graph.
GAT is the most representative method of Non-Local Neural Networks family, where the updating of each node is based on a weighted sum of the node attributes of its neighbors.

Besides neighborhood, nodes in the real-world are also with rich side information and attributes, which could be aggregated in the same manner.
Some researchers attempt to model network structure and side information simultaneously \cite{DBLP:conf/ijcai/YangLZSC15,DBLP:conf/acl/TuLLS17,DBLP:conf/ijcai/PanWZZW16,DBLP:conf/nips/HamiltonYL17}.
These jointly-learning approaches mainly integrate the topological and side information based on framework like Skip-Gram, or matrix factorization equivalently (the equivalence is discussed in \cite{DBLP:conf/wsdm/QiuDMLWT18}).

As for our proposed CNE framework, the biggest difference with the existing compositional methods lies in the input of the model.
In the existing methods, (1) all or part of the input components come from the neighborhoods (\textit{i.e.} GraphSAGE \cite{DBLP:conf/nips/HamiltonYL17}, GCN \cite{DBLP:journals/corr/KipfW16}) and (2) non-compositional encoders are preserved for topology modelling (\textit{i.e.} CANE \cite{DBLP:conf/nips/HamiltonYL17}, TriDNR\cite{DBLP:conf/ijcai/PanWZZW16}).
CNE only takes the components from the target node as input and node ID is completely abandoned.
This difference makes CNE outperform the existing methods when inferring the embeddings of the unseen nodes which are not connected to the network in training phase.
Only the encoder of CNE leverages the same amount of information in the training and inference phase.
The existing methods, on the contrary, could leverage partial or even none information in the inference.

\section{Compositional Network Embedding}

\tikzset{
  emb/.style = {rectangle, fill=flamingo, minimum width=5em, minimum height=1.5em},
  tit/.style = {text depth=.25ex, text height=1.5ex, inner sep=0, outer sep=0},
  barbox/.style={rectangle, inner sep=0, outer sep=0, fill=padua, anchor=south, minimum width=2mm},
  bemb/.style = {rectangle, fill=flamingo, minimum width=1em, minimum height=2em},
  bbarbox/.style={rectangle, inner sep=0, outer sep=0, fill=shakespeare, anchor=south, minimum width=2mm},
  dec/.style = {rectangle, fill=my_sin, minimum width=1em, minimum height=2em},
  tcv/.style = {rectangle, fill=carmine_pink, minimum width=1em, minimum height=2em},
  comp_func/.style={rectangle, draw, anchor=south, minimum height=1em, minimum width=6em},
  g_node/.style={circle, inner sep=0, outer sep=0, fill=platinum, anchor=south, minimum width=6mm},
  feature/.style={
       rectangle split,
       rectangle split horizontal=true,
       rectangle split draw splits=false,
       rectangle split parts=7,
       rectangle split part fill={red!30, blue!20, macaroni_and_cheese, white, persian_green, silver, bittersweet},
       draw=black, %
       minimum height=2em,
       minimum width=3em,
       inner sep=2pt,
       },
  feature2/.style={
       rectangle split,
       rectangle split horizontal=true,
       rectangle split draw splits=false,
       rectangle split parts=7,
       rectangle split part fill={gallery, bittersweet, free_speech_aquamarine, white, macaroni_and_cheese, celestial_blue, carnation},
       draw=black, %
       minimum height=2em,
       minimum width=3em,
       inner sep=2pt,
       },
  feature3/.style={
       rectangle split,
       rectangle split horizontal=true,
       rectangle split draw splits=false,
       rectangle split parts=7,
       rectangle split part fill={aluminum, fern, jelly_bean, white, jet_stream, sushi, egg_shell},
       draw=black, %
       minimum height=2em,
       minimum width=3em,
       inner sep=2pt,
       },
}

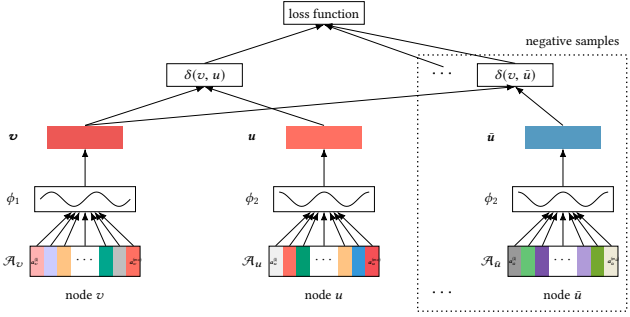
\begin{figure}
  \centering
  \resizebox{\linewidth}{!}{
  \begin{tikzpicture}[]
  \tikzstyle{every node}=[font=\small]

    \node[feature, node distance=6cm, yshift=-0.5cm] (f_a) at (0, 0) {
    \nodepart{four} $\cdots$
    };
    \node[left of=f_a, node distance=1.5cm] (a_a) {$\mathcal{A}_v$};
    \node[left of=f_a, node distance=1.05cm, scale=0.45] (a_a1) {$\bm{a}_v^{(\!1\!)}$};
    \node[right of=f_a, node distance=1.06cm, scale=0.45] (a_an) {$\bm{a}_v^{(\!n_v\!)}$};
    
    \node[below of=f_a, node distance=0.7cm] (i_a) {node $v$};
    
    \node[comp_func, above of=f_a, node distance=1.3cm] (cf_a) {\tikz \draw[scale=0.15,domain=-6.282:6.282,smooth,variable=\t]
     plot (\t,{sin(\t r)});};
     
    \node[left of=cf_a, node distance=1.5cm] (a_ca) {$\phi_1$};

    \draw[-Latex] (f_a) -- (cf_a) ;
    
    \draw[-Latex] (f_a.one north) -- (cf_a) ;
    \draw[-Latex] (f_a.two north) -- (cf_a) ;
    \draw[-Latex] (f_a.three north) -- (cf_a) ;
    \draw[-Latex] (f_a.five north) -- (cf_a) ;
    \draw[-Latex] (f_a.six north) -- (cf_a) ;
    \draw[-Latex] (f_a.seven north) -- (cf_a) ;
    
    \node[emb, above of=cf_a, node distance=1.3cm] (ne_a) {};
    \node[left of=ne_a, node distance=1.5cm] (ae_a) {$\bm{v}$};
    
    \node[feature2, right of=f_a, node distance=5cm] (f_b) {
    \nodepart{four} $\cdots$
     };
     
    \node[left of=f_b, node distance=1.05cm, scale=0.45] (a_b1) {$\bm{a}_u^{(\!1\!)}$};
    \node[right of=f_b, node distance=1.06cm, scale=0.45] (a_bn) {$\bm{a}_u^{(\!n_v\!)}$};
    
    \node[below of=f_b, node distance=0.7cm] (i_b) {node $u$};
    \node[left of=f_b, node distance=1.5cm] (a_b) {$\mathcal{A}_u$};
    
    \node[comp_func, above of=f_b, node distance=1.3cm] (cf_b) {\tikz \draw[scale=0.15,domain=-6.282:6.282,smooth,variable=\t]
     plot (\t,{cos(\t r)});};
    
    \node[left of=cf_b, node distance=1.5cm] (a_cb) {$\phi_2$};
    
    \draw[-Latex] (f_b) -- (cf_b) ;
    \draw[-Latex] (f_b.one north) -- (cf_b) ;
    \draw[-Latex] (f_b.two north) -- (cf_b) ;
    \draw[-Latex] (f_b.three north) -- (cf_b) ;
    \draw[-Latex] (f_b.five north) -- (cf_b) ;
    \draw[-Latex] (f_b.six north) -- (cf_b) ;
    \draw[-Latex] (f_b.seven north) -- (cf_b) ;
    
    \node[rectangle, fill=bittersweet, minimum width=5em, minimum height=1.5em, above of=cf_b, node distance=1.3cm] (ne_b) {};
    \node[left of=ne_b, node distance=1.5cm] (ae_b) {$\bm{u}$};
    
    \node[rectangle, draw, minimum width=5em, minimum height=1.5em, above of=ne_a, node distance=1.3cm, xshift=2.5cm] (s_1) {$\delta(v,u)$};
    
    \draw[-Latex] (cf_a) -- (ne_a) ;
    \draw[-Latex] (cf_b) -- (ne_b) ;
    \draw[-Latex] (ne_a.north) -- (s_1.south) ;
    \draw[-Latex] (ne_b.north) -- (s_1.south) ;

    \node[feature3, right of=f_b, node distance=5cm] (f_c) {
    \nodepart{four} $\cdots$
     };
     
     \node[left of=f_c, node distance=1.05cm, scale=0.45] (a_c1) {$\bm{a}_{\bar{u}}^{(\!1\!)}$};
    \node[right of=f_c, node distance=1.06cm, scale=0.45] (a_cn) {$\bm{a}_{\bar{u}}^{(\!n_{\bar{u}}\!)}$};
     
    \node[right of=i_b, node distance=2.5cm] (i_n) {\Large $\cdots$};
    
    \node[below of=f_c, node distance=0.7cm] (i_c) {node $\bar{u}$};
    \node[left of=f_c, node distance=1.5cm] (a_c) {$\mathcal{A}_{\bar{u}}$};
    
    \node[comp_func, above of=f_c, node distance=1.3cm] (cf_c) {\tikz \draw[scale=0.15,domain=-6.282:6.282,smooth,variable=\t]
     plot (\t,{cos(\t r)});};
    
    \node[left of=cf_c, node distance=1.5cm] (a_cc) {$\phi_2$};

    \draw[-Latex] (f_c) -- (cf_c) ;
    \draw[-Latex] (f_c.one north) -- (cf_c) ;
    \draw[-Latex] (f_c.two north) -- (cf_c) ;
    \draw[-Latex] (f_c.three north) -- (cf_c) ;
    \draw[-Latex] (f_c.five north) -- (cf_c) ;
    \draw[-Latex] (f_c.six north) -- (cf_c) ;
    \draw[-Latex] (f_c.seven north) -- (cf_c) ;
    
    \node[rectangle, fill=shakespeare, minimum width=5em, minimum height=1.5em, above of=cf_c, node distance=1.3cm] (ne_c) {};
    \node[left of=ne_c, node distance=1.5cm] (ae_c) {$\bar{\bm{u}}$};
    
    \node[rectangle, draw, minimum width=5em, minimum height=1.5em, above of=ne_c, node distance=1.3cm, xshift=-1cm] (s_2) {$\delta(v,\bar{u})$};
    \draw[-Latex] (cf_c) -- (ne_c) ;
    \draw[-Latex] (ne_a.north) -- (s_2.south) ;
    \draw[-Latex] (ne_c.north) -- (s_2.south) ;
    
    \node[rectangle, draw, minimum width=5em, minimum height=1.5em, above of=ne_b, node distance=2.6cm] (loss) {loss function};
    
    \draw[-Latex] (s_1.north) -- (loss.south) ;
    \draw[-Latex] (s_2.north) -- (loss.south) ;
    
    \node[left of=s_2, node distance=1.5cm] (s_n) {\Large $\cdots$};
    \draw[-Latex] (s_n.north) -- (loss.south) ;
    
    \node[above of=s_2, node distance=0.7cm, xshift=1.2cm] (ns) {negative samples};
    
    \node[draw, thick, dotted, inner sep = 5pt, fit=(s_n) (s_2) (i_n) (f_c)] {};

  \end{tikzpicture}
  }
  \caption{Compositional network embedding framework. For each node $v$ and its neighbor $u$, we randomly sample $K$ negative node $\bar{u}$. The objective of our framework is to distinguish the positive node $u$ from the negative node $\bar{u}$ using the embeddings derived from their internal attributes.}
  \label{fig:cne}
\end{figure}

In this section, we will present our idea of compositional network embedding (CNE) framework. %
The core idea behind our approach is that we learn how to derive the node embeddings from the embeddings of the features they carried.
As Fig.~\ref{fig:cne} illustrated, CNE framework contains two key parts:
\begin{enumerate*}[label=(\roman*)]
\item composition function that generates node embeddings from their internal attributes;
\item unsupervised graph-based loss function for learning the parameters of composition function and attribute embeddings.    
\end{enumerate*}
Here, without loss of generality, we take text features as an example to introduce the framework for the sake of simplicity.

\subsection{Embedding Composition}

In this subsection, we introduce the embedding composition procedure in our framework.
As Fig.~\ref{fig:cne} illustrated, the embedding generation process for a node $v_i$ is very simple and straightforward, applying the composition function $\phi$ to the node $v_i$'s features embeddings $\mathcal{A}_i {=} [\bm{a}_i^{(1)},\dots,\bm{a}_i^{(n_i)}]$ as: %
\begin{equation*}
	\bm{v}_i = \phi(\mathcal{A}_i) =\phi \Bigl(\bm{a}_i^{(1)},\ldots, \bm{a}_i^{(n_i)} \Bigr)
\end{equation*}
where, $\bm{a}_i^{(j)} \in \mathbb{R}^d$ is the embedding of node $v_i$'s attributes $a_i^{(j)}$, $d$ is the dimensionality of $\bm{a}_i^{(j)}$, $n_i$ is the length of $v_i$'s features.
Here, we model the composition function $\phi$ as a neural network.
In this context, the composition function is equivalent to an encoder.
Thus, we also use \textit{encoder} to refer to the \textit{composition function} later.
It is worth emphasizing that feature $a$ and its embedding $\bm{a}$ are shared across all nodes in the network.
In this way, after the model has been trained and the parameters are fixed, we can generate the embedding for an unseen node by feeding these shared feature embeddings through the composition function $\phi$.

\subsection{Composition Function (Encoder)} 

CNE is a quite general framework for network embedding.
Here, the ``generality'' is reflected in the choices for feature design and composition function design.
For example, we can use text (\textit{e.g.}, product title) as the feature for the networks like product network, or image as the feature for Flickr image relationships. %
According to the different features, we can design the corresponding encoders, from simply concatenate, mean, and sum operator to complex models like GRU \cite{DBLP:journals/corr/ChungGCB14} and CNN \cite{Lecun:Gradient}. %

In this paper, without loss of generality, we mainly focus on the network with text features to introduce the framework for the sake of simplicity.
Here, we use an RNN encoder with GRU as composition function $\phi$ to encode text features. 
The last hidden state $\bm{h}_{v_i}^{(n_i)}$ are used as the representations of the node features $\mathcal{A}_i$, which we also treat as the node embedding of node $v_i$.
At each time step $t$, the GRU is parameterized as (omitting the subscript for notational convenience):
\begin{equation*}
\begin{aligned}
  \bm{r}^{(t)} &= \sigma\bigl(\bm{W}_r \bm{a}^{(t)} + \bm{U}_r \bm{h}^{(t-1)}\bigr) \\
    \bm{z}^{(t)} &= \sigma\bigl(\bm{W}_z \bm{a}^{(t)} + \bm{U}_z \bm{h}^{(t-1)}\bigr) \\
    \tilde{\bm{h}}^{(t)} &= \tanh\bigl(\bm{Wa}^{(t)} + \bm{U}(\bm{r}^{(t)}\odot\bm{h}^{(t-1)})\bigr) \\
    \bm{h}^{(t)} &= \bigl(1-\bm{z}^{(t)}\bigr)\odot\bm{h}^{(t-1)} + \bm{z}^{(t)}\odot\tilde{\bm{h}}^{(t)}
\end{aligned}
\end{equation*}
where $\sigma$ is the sigmoid function, $\odot$ is element-wise multiplication, $\bm{r}^{(t)}$ is \textit{reset gate}, $\bm{z}^{(t)}$ is \textit{update gate}, all the non-linear operations are computed element-wise, and weight matrices like $\bm{W}_r, \bm{W}, $ and $\bm{U}$ are all learnable parameters.
While we use GRU here, any other types of encoder can be used so long as we can back-propagate through it.

\subsection{Learning The Parameters of CNE}

In order to learn useful and predictive embeddings in an unsupervised way, we apply a graph-based loss to the output node embeddings, and tune the parameters via stochastic gradient descent.
The graph-based loss function encourages nearby nodes to have similar embeddings, while enforcing the embeddings of disparate nodes are highly dissimilar.
Specifically, CNE is trained end-to-end in a \textit{siamese} framework, as illustrated in Fig.~\ref{fig:cne}.
Formally, for node $v$ and its neighbor $u{\in} \mathcal{N}(v)$ ($\mathcal{N}(v)$ is the neighbor set for $v$), we define the max-margin (\textit{i.e.}, hinge) loss function \cite{DBLP:journals/jmlr/CollobertWBKKK11,lazaridou-dinu-baroni:2015:ACL-IJCNLP} as:
\begin{equation}
\mathcal{L}(v, u) = \sum_{k=1}^{K} \max\bigl(0, m - \delta(v, u) + \delta(v, \bar{u}_k)\bigr)
\label{eq:loss}
\end{equation}
where $\bar{u}_k$ is a negative sample randomly sampled from the whole node set $\mathcal{V}$, $K$ is the number of negative samples; $m$ is the margin between the the positive node pairs and the negative node pairs, usually set as 1; $\delta$ is the score function to measure the similarity between two nodes, we define as:
\begin{equation*}
	\delta(v, u) = \cos(\bm{v}, \bm{u}) = \cos\Bigl(\phi_1(\mathcal{A}_v), \phi_2(\mathcal{A}_u)\Bigr)
\end{equation*}
where, $\phi_1$ and $\phi_2$ are encoders for node $v$ and $u$ respectively.

Instead of getting the vector $\bm{v}$ (and $\bm{u}$) via an embedding look-up, CNE generates them by combining the feature embedding carried with node $v$ (and $u$).
The compositional embedding makes it possible to interact between network topology and feature similarity.

Intuitively, the goal of the max-margin objective is to rank the correct neighbor $u$ of node $v$ higher than any other random node $\bar{u}$ with a margin $m$.
Importantly, other loss functions (\textit{e.g.}, likelihood objective in DeepWalk) are also valid in our framework.
Here, we choose hinge loss for its better performance in our experiments.
It is worth noting that CNE can also be trained in a supervised manner, by simply replacing (or augmenting) the unsupervised loss (Eq.~\ref{eq:loss}) with a task-specific objective.

\subsubsection{Neighborhood definition.} As mentioned above in the loss function, the neighborhood definition is a key part in the training stage.
In this paper, we define the neighborhood based on random walks as the same in DeepWalk for its effectiveness and efficiency.
Specifically, we first start truncated random walks with length $l$ at each node.
After that, the \textit{neighbors} of node $v$ can be defined as the set of nodes within a window size $w$ in each random walk sequence.
This sample strategy makes the node embedding similarity relevant to the geodesic distance.

\subsection{CNE for Various Kinds of Networks}

As we presented above, CNE is a general framework for network representation learning. %
In CNE, node embeddings are mainly determined by the nodes' attributes and the encoders (\textit{i.e.}, models and parameters).
Here, we will discuss how to apply CNE to various kinds of networks by adjusting the setting of node attributes and encoders accordingly.

\subsubsection{Directed network.} In this case, edge direction is supposed to be preserved.
For this purpose, we can learn node $v$'s encoder $\phi_1$ and its neighbor $u$'s encoder $\phi_2$ using different parameters respectively.
In this way, $\delta(v, u) \neq \delta(u, v)$ since $\phi_1(\mathcal{A}_v) \neq \phi_2(\mathcal{A}_v)$. %

\subsubsection{Heterogeneous network with multiple node types.} 
In this case, different types of nodes usually carry with features from different fields.
Item attributes, for instance, seem to be completely different from the user demographic characteristics in recommendation field.
CNE is naturally capable of mapping different types of nodes to the same low-dimension space. %
We can simply employ different models as encoders according to the types of nodes.
For example, we can use RNN to encode nodes with text features, use CNN to encode nodes with images, or use more complex network for nodes with features from various fields.%

\subsubsection{Heterogeneous network with multiple edge types.}
CNE models multiple edge types by learning several pairs of encoders but sharing the same feature embeddings.
For example, the social network users have follow and retweet relationships. 
We build four encoders altogether, two for modeling follow relationship and the other two for the retweet relationship.
CNE is more like a multi-task model in this case.

\section{Experiments}
In this section, we test the performance on four sub-tasks of link prediction, to demonstrate the model's ability and flexibility on both homogeneous and heterogeneous network.

\subsection{Baselines}

To verify the effectiveness of our proposed model, we compare it with several strong baselines, including:

\textbf{SGNS} \cite{Mikolov:2013:DRW:2999792.2999959}: It represents the node as the sum of the corresponding word embeddings which are learned by applying SGNS to the text sequences generated from node features.

\textbf{DeepWalk} \cite{DBLP:conf/kdd/PerozziAS14}: It derives the node embeddings by combining random walks and Skip-Gram model. 

\textbf{CANE} \cite{DBLP:conf/acl/TuLLS17}: It learns context-aware embeddings for nodes with mutual attention mechanism and associated text information. 

\textbf{TriDNR} \cite{DBLP:conf/ijcai/PanWZZW16}: It learns node embeddings by  jointly modeling the network structure, node-content correlation, and label-content correspondence.

\textbf{GraphSAGE} \cite{DBLP:conf/nips/HamiltonYL17}: It is an inductive model that generates node embeddings by aggregating features from a node's neighborhood.

\subsection{Parameters Setting}

For all baselines, we used the implementation released by the original authors.
We implement CNE using \texttt{TensorFlow}\footnote{\url{https://www.tensorflow.org}}.
For fair comparison, we train these baselines with the same random-walk relevant settings as our proposed model.
The node embedding dimension of all models is set to 512.
For all models that leverage text features, we build a vocabulary of top \num{40000} words and learn the word embeddings from scratch.
For GraphSAGE, we use mean operator as the aggregator and train word vectors trained by SGNS.
For models based on random walks (\textit{e.g.}, DeepWalk, TriDNR, and CNE), we set the length of truncated random walks as $l {=} 20$, set window size $w {=} 2$, and randomly sample $K {=} 4$ negative nodes for each positive node pair. %
For GRU encoder in CNE, we use 256-dimensional word embeddings and 512-dimensional hidden states for GRU units without any attention mechanism.
We train CNE using Adam \cite{Diederik:Adam} with initial learning rate of 8e-4 and batch size of 256.

\subsection{Task and Evaluation Metrics}

In this paper, we use the link prediction (LP) task to evaluate the ability of our proposed model under different settings. %
We randomly remove a portion of existing edges from the network and use the left network to train each network embedding model.
For testing, we randomly choose one thousand nodes from the network and use the learned node embeddings to predict the unobserved links.
Unlike previous works solve the link prediction as a binary classification problem, we adopt the experimental settings in \cite{DBLP:conf/kdd/WangC016,Peng:survey:TKDE} since it is more practical and reasonable.
This is because, based on the learned embeddings, the similarity between nodes can be easily estimated, \textit{e.g.}, by the cosine similarity or the inner product. 
Intuitively, a larger similarity implies that the two nodes may have a higher propensity to be linked.
In this way, we can employ Precision$@k$ and Recall$@k$ to evaluate the link prediction performance:
\begin{equation*}
\begin{aligned}
  \mathrm{Precision}@k &= \frac{\#(\text{real neighbors} \cap \text{top } k \text{ candidates})}{k} \\
\mathrm{Recall}@k &= \frac{\#(\text{real neighbors} \cap \text{top } k \text{ candidates})}{\#(\text{real neighbors})}. \label{eq:recall}
\end{aligned}
\end{equation*}
We calculates Precision@$k$ and Recall@$k$ score for each node and report the results as the average scores of all test nodes.
It is worth noting that ``top $k$ candidates'' are truncated from the whole list ranked on the candidate set.
Here, instead of using a small candidate set built from test set as in \cite{DBLP:conf/kdd/WangC016,Peng:survey:TKDE}, we calculate the rank list base on the whole node set $\mathcal{V}$. 
This is a more difficult, but practical and reasonable setting.

\begin{table}
\caption{Recall of different training set size (Task 1).\label{T1P}}
\begin{adjustbox}{max width=\linewidth}
\begin{tabular}{@{}l cc cc cc cc cc@{}}
\toprule
         & \multicolumn{2}{c}{10\%}                   & \multicolumn{2}{c}{30\%}                   & \multicolumn{2}{c}{50\%} & \multicolumn{2}{c}{70\%}                    & \multicolumn{2}{c}{90\%}                    \\
\cmidrule(lr){2-3} \cmidrule(lr){4-5} \cmidrule(lr){6-7} \cmidrule(lr){8-9} \cmidrule(l){10-11}
Method   & R@10         & R@100          & R@10         & R@100          & R@10     & R@100   & R@10         & R@100          & R@10        & R@100          \\ \midrule
SGNS & 0.054        & 0.134          & 0.056          & 0.136          & 0.055       & 0.136   & 0.055          & 0.138           & 0.056    & 0.137           \\
DeepWalk & 0.103        & 0.389          & 0.102       & 0.430          & 0.112    & 0.456   & 0.114       & 0.462          & 0.116        & 0.469 \\
TriDNR   & 0.091          & 0.243          & 0.095         & 0.278          & 0.102    & 0.299   & 0.101& 0.309          & 0.104  & 0.329      \\
CANE     & 0.140      & 0.473          & 0.138          & 0.478            &   0.144     & 0.484   &     0.138       &     0.477     & 0.139              & 0.483            \\
GraphSAGE & 0.088    & 0.308          & 0.101       & 0.396            &   0.113    & 0.437  &     0.112      &     0.446     & 0.103    & 0.436           \\
CNE      & \textbf{0.154} &  \textbf{0.547} & \textbf{0.152} &  \textbf{0.551} &   \textbf{0.157}   & \textbf{0.564}     & \textbf{0.162} & \textbf{0.585} & \textbf{0.176} & \textbf{0.622} \\ \bottomrule
\end{tabular}
\end{adjustbox}
\end{table}

\begin{table}
\caption{Precision of different training set size (Task 1). \label{T1R}}
\begin{adjustbox}{max width=\linewidth}
\begin{tabular}{@{}l cc cc cc cc cc@{}}
\toprule
             & \multicolumn{2}{c}{10\%}                   & \multicolumn{2}{c}{30\%}                   & \multicolumn{2}{c}{50\%} & \multicolumn{2}{c}{70\%}                    & \multicolumn{2}{c}{90\%}                        \\ 
\cmidrule(lr){2-3} \cmidrule(lr){4-5} \cmidrule(lr){6-7} \cmidrule(lr){8-9} \cmidrule(l){10-11}
Method       & P@10           & P@100          & P@10           & P@100          & P@10      & P@100     & P@10         & P@100          & P@10     & P@100          \\ \midrule
SGNS     & 0.131             & 0.036          & 0.132                & 0.036          & 0.131       & 0.036     & 0.133             & 0.037          & 0.134          & 0.037           \\
DeepWalk     & 0.208      & 0.105          & 0.220           & 0.117          & 0.237       & 0.123     & 0.237        & 0.123          & 0.244  & 0.125          \\
TriDNR       & 0.190            & 0.062          & 0.203          & 0.072          & 0.215      & 0.078     & 0.213          & 0.082           &0.219 & 0.086         \\
CANE         & 0.315            & 0.130          & 0.320             & 0.133          &  0.328      &0.133     &      0.315      &    0.130      &0.316          &0.131           \\
GraphSAGE         & 0.216         & 0.086          & 0.252              & 0.111          &  0.263     &0.120     &   0.261         &    0.123      &0.236                  &0.120           \\
CNE & \textbf{0.374}  & \textbf{0.155} & \textbf{0.369} &  \textbf{0.157} & \textbf{0.389 } & \textbf{0.161} & \textbf{0.390} &  \textbf{0.166} & \textbf{0.414} &  \textbf{0.174} \\ \bottomrule
\end{tabular}
\end{adjustbox}
\end{table}

\begin{table}[!tb]
\centering
\caption{Precision of unseen test nodes (Task 2).\label{T2}}
\begin{adjustbox}{max width=\linewidth}
\begin{tabular}{@{}l cc cc cc @{}}
\toprule
                      & \multicolumn{2}{c}{10\%}                    & \multicolumn{2}{c}{30\% }                    & \multicolumn{2}{c}{50\% }\\ 
\cmidrule(lr){2-3} \cmidrule(lr){4-5} \cmidrule(lr){6-7}
Method    & P@10  & P@100          & P@10  & P@100          & P@10  & P@100         \\ \midrule
SGNS   & 0.016 & 0.003 & 0.016& 0.003 & 0.016 & 0.003 \\
TriDNR & 0.019 & 0.003 & 0.021 & 0.004 & 0.030 & 0.005 \\
CNE    & \textbf{0.034} &\textbf{0.008} & \textbf{0.039} &  \textbf{0.009} & \textbf{0.042} &  \textbf{0.009} \\ \bottomrule
\end{tabular}
\end{adjustbox}
\end{table}

\subsection{Task 1: LP on Homogeneous Network} %

This task is to validate that CNE possesses higher representation power although the network is incomplete.

We use Amazon Baby category dataset provided by \cite{DBLP:conf/www/HeM16}, consisting of \num{71317} item's metadata.
The homogeneous network is constructed from \textit{co-view} relations, resulting in an item graph with \num{47185} nodes and \num{1166828} edges.
The portion of removed edges is varied between $10\%$ and $90\%$.
We treat the product title as node feature.
It is a short description of the product usually, \textit{e.g.}, \textit{Lifefactory 4oz BPA Free Glass Baby Bottles}. %

\textbf{Results } As shown in Table \ref{T1P} and Table \ref{T1R}, CNE achieves much better scores than all the baselines.
Note that DeepWalk and CANE require that all nodes in the graph are present during training of the embeddings.
For fairness of comparison, we only use the nodes presented in the training network to construct the test set.

\textbf{Importance of network topology. } %
It is a little surprising that DeepWalk is a strong baseline in our experiments, indicating that structure provides rich information for link prediction.
SGNS is based on node features solely, getting the word similarity only from word co-occurrence statistics.
The gaps between CNE and SGNS suggest that network topology can improve the attribute embeddings for network embedding. 
This result is in consistent with the finding in \cite{kenter-borisov-derijke:2016:P16-1}.

\textbf{Importance of jointly-training topology and attributes. } %
As node attributes can alleviate the data sparsity problem to some extent, they do enhance the structure-based embeddings when being utilized in a reasonable way. 
Comparing with DeepWalk, the results show that models leveraged nodes' internal attributes (\textit{e.g.}, CANE and CNE) can achieve better performances, especially on small training set.
Although these methods take advantage of the nodes' text attributes, they choose different ways.
TriDNR and CANE all optimize the attribute embeddings indirectly through the node embeddings (The former uses a shadow model and the latter uses deep CNN); while GraphSAGE uses a fixed pre-trained attribute embeddings.
The results show that our proposed CNE outperforms these baselines with a significant gap.
This indicates that directly optimizing the node attribute embeddings and the encoders in an end-to-end way can act as a powerful form of regularization to improve the performances.
In addition, comparing results from different groups, it is easy to find that CNE is more stable than other baselines, improving consistently along with the increase of the training set.

\subsection{Task 2: LP for Unseen Nodes}

This task is designed to test the ability of CNE on generating the embedding for unseen nodes.
We use the same experimental setting as Task 1 except for the choice of test set.
In this task, we focus on the nodes that did not appeared in the training set.
GraphSAGE, DeepWalk, and CANE are not available for this task as they can not calculate the node embeddings without nodes structural information.
Therefore, only the results of the rest baselines are presented in Table~\ref{T2}.
In this task, we only conduct the experiments on networks with portion of training edges varied from $10\%$ to $50\%$ since keeping more edges will not be able to generate enough unseen nodes for test.
In addition, we will only present Precision@$k$ for the following experiments due to the page limitation.

\textbf{Results } Table~\ref{T2} includes the comparison between different baselines.
At most of time, CNE is still the best one among them.
CNE encodes the network structure by feature embeddings and encoders, while TriDNR separately models the structural and semantic embeddings.
When TriDNR is without structural information, the representation power of the semantic embeddings is too weak to accomplish link prediction task.
CNE on the contrary does not suffer from such trouble as long as it captures sufficient connections between network structure and semantic distances.
Meanwhile, in the training of TriDNR, the word embeddings interact with network topology.
Therefore, it outperforms SGNS, whose word embeddings is based on the context (word co-occurrence statistics).
Note that Task 2 is much harder than Task 1, as the unseen nodes have quite few edges.

\subsection{Task 3: LP on Multi-edge-type Network}

This task is designed to show that CNE is easily extensible and able to jointly model several node relations.
We still use the Amazon baby category dataset and construct the network of two kinds of node relations, \textit{co-view} and \textit{buy-after-view}.
The \textit{co-view} network consists of \num{47185} nodes and \num{1166828} edges, the \textit{buy-after-view} network consists of \num{44078} nodes and \num{111473} edges.
The result of \textit{buy-after-view} is presented in Table~\ref{T3}.

\begin{table}
\caption{Precision on the \textit{buy-after-view} network (Task 3).}
\label{T3}
\begin{adjustbox}{max width=\linewidth}
\begin{tabular}{@{}l cc cc cc cc@{}}
\toprule
 & \multicolumn{2}{c}{20\%}  & \multicolumn{2}{c}{40\%}  & \multicolumn{2}{c}{60\%}   & \multicolumn{2}{c}{80\%}                   \\ 
\cmidrule(lr){2-3} \cmidrule(lr){4-5} \cmidrule(lr){6-7} \cmidrule(l){8-9} 
   & P@10             & P@100            & P@10             &  P@100            & P@10        & P@100       & P@10           & P@100           \\
\midrule
SGNS      & 0.033               & 0.007          & 0.034         & 0.007          & 0.035            & 0.007          & 0.036            & 0.007          \\
DeepWalk  & 0.093           & 0.019          & 0.099              & 0.023          & 0.095           & 0.023          & 0.090           & 0.022          \\
CANE      & 0.080        & 0.017          & 0.092          &  0.019          & 0.091        & 0.019          & 0.091           & 0.019          \\
TriDNR    & 0.065       & 0.013          & 0.068              & 0.014          & 0.078           & 0.016          & 0.078              & 0.017          \\
GraphSAGE & 0.056            & 0.012          & 0.063             & 0.014          & 0.067      & 0.016          & 0.068       & 0.016          \\
CNE       & 0.081          & 0.019          & 0.085               & 0.022          & 0.083            & 0.024          & 0.082           & 0.030          \\
CNE$_{\mathrm{MUL}}$    & \textbf{0.120}  & \textbf{0.022} & \textbf{0.128}  & \textbf{0.022} & \textbf{0.134} &  \textbf{0.027} & \textbf{0.136}& \textbf{0.033}        \\
\bottomrule
\end{tabular}
\end{adjustbox}
\end{table}

The \textit{buy-after-view} network is so sparse that each node has 5 edges on average.
Modeling this sparse network by CNE is not a wise choice as the model fails to be well trained with such small amount of edges.
We address this problem by multi-task learning, building two encoders and loss functions for \textit{co-view} and \textit{buy-after-view} networks respectively. 
It is observed that for a certain node, at most of time, the \textit{buy-after-view} neighbors compose a small subset of \textit{co-view} neighbors.
The \textit{co-view} encoder and loss can be seen as the guidance of the \textit{buy-after-view} training.

\textbf{Results } As shown in Table \ref{T3}, CNE$_{\mathrm{mul}}$ has much better performance than all the baselines, where the former one is the multi-task learning CNE and the latter one models the \textit{buy-after-view} network directly.
DeepWalk also has competitive performances as CNE.
On the \textit{buy-after-view} network, other baselines tend to be under-fitting.
DeepWalk has fewer parameters to learn, so that the impact of edge sparsity is weaker on DeepWalk.
CNE has similar advantages as it learns the feature embeddings and aggregators, which are shared among all the nodes.
GraphSAGE is not suitable for the \textit{rare-neighbors} situation as the node embedding in GraphSAGE is mainly based on modelling node neighbors. 

\subsection{Task 4: LP on Multi-node-type Network}

This task is designed to validate that CNE is pretty flexible to model the network of multiple node types.
In this experiment, we collect users' behavior sequences from a popular online shopping website.
The task is to predict a user's subsequent behavior ($n$+1) given the past $n$ behaviors.
Baselines need a homogeneous network and we construct a item graph by connecting two items if they are viewed in the same session.
The network consists of \num{8744} nodes and \num{29976} edges.
All the baselines provide a node embedding for each item and the behavior sequence is represented by a \textit{record} vector by adding $n$ items embeddings up.
Meanwhile, we construct a heterogeneous network of two kinds of nodes, user and item.
Each user node is associated with $n$ item viewing records within a session.
CNE models the user node by $n$ GRU encoders and produce the user embedding (\textit{record} vector) by adding up the last hidden layer of $n$ encoders.
In our experiments, we let $n$ equal 4.

Compared with the above tasks, Task 4 is more challenging as each record only corresponds to one correct answer, the most subsequent item.
We calculate the cosine similarity between the candidate items' vector and the \textit{record} vector, ranking the candidates in the descending order of cosine similarity.
Note that the record products are filtered out from the candidate products.
We test 1000 users' records and count the positions of the correct answers in the 1000 ordered lists.

\begin{figure}
\pgfplotsset{
axis background/.style={fill=gallery},
grid=both,
  xtick pos=left,
  ytick pos=left,
  tick style={
    major grid style={style=white,line width=.5pt},
    minor grid style={style=bgc, line width=.1pt},
    draw=none
    },
  ytick={200, 400},
  every x tick label/.append style={font=\tiny, yshift=2pt},
  every y tick label/.append style={rotate=90, font=\fontsize{1pt}{1pt}\selectfont, yshift=-2pt},
  ymajorgrids,
	major grid style={draw=white},
	y axis line style={opacity=0},
	tickwidth=0pt,
  title style={yshift=-5ex, font=\small},
  y label style={yshift=-5pt},
  minor tick num=1,
  height=2.5cm,
  width=1.2\linewidth,
  ymin=0, ymax=600
}
\centering
\begin{tikzpicture}[]
\node (t) at (0,0) {};	
\end{tikzpicture}
\resizebox{\linewidth}{!}{
\begin{tikzpicture}
	\begin{axis}[
		smooth,
		area style,
		enlarge x limits=false,
		xticklabels={0-2, 2-4, 4-8, 8-16, 16-32, 32-64, 64-128, 128-256, 256-512, 512-1024},
	    xtick={1,2,3,4,5,6,7,8,9,10},
	    ylabel={\scriptsize Counting},
	    title={DeepWalk},
	    ]
	\addplot[color=puerto_rico, fill=puerto_rico] coordinates
		{(1, 130) (2, 531) (3, 237) (4, 46) (5, 7) (6, 3) (7, 2) (8, 5) (9, 7) (10, 0)}
		\closedcycle;
	\end{axis}
\end{tikzpicture}
}
\resizebox{\linewidth}{!}{
\begin{tikzpicture}
	\begin{axis}[
		smooth,
		area style,
		enlarge x limits=false,
		xticklabels={0-2, 2-4, 4-8, 8-16, 16-32, 32-64, 64-128, 128-256, 256-512, 512-1024},
	    xtick={1,2,3,4,5,6,7,8,9,10},
	    ylabel={\scriptsize Counting},
	    title={CANE},
	    ]
	\addplot[color=biscay, fill=biscay] coordinates
		{(1, 6) (2, 7) (3, 15) (4, 31) (5, 62) (6, 116) (7, 194) (8, 287) (9, 247) (10, 3)}
		\closedcycle;
	\end{axis}
\end{tikzpicture}
}
\resizebox{\linewidth}{!}{
\begin{tikzpicture}
	\begin{axis}[
		smooth,
		area style,
		enlarge x limits=false,
		xticklabels={0-2, 2-4, 4-8, 8-16, 16-32, 32-64, 64-128, 128-256, 256-512, 512-1024},
	    xtick={1,2,3,4,5,6,7,8,9,10},
	    ylabel={\scriptsize Counting},
	    title={TriDNR},
	    ]
	\addplot[color=celestial_blue, fill=celestial_blue] coordinates
		{(1, 19) (2, 151) (3, 122) (4, 106) (5, 103) (6, 96) (7, 100) (8, 111) (9, 157) (10, 3)}
		\closedcycle;
	\end{axis}
\end{tikzpicture}
}
\resizebox{\linewidth}{!}{
\begin{tikzpicture}
	\begin{axis}[
		smooth,
		area style,
		enlarge x limits=false,
		xticklabels={0-2, 2-4, 4-8, 8-16, 16-32, 32-64, 64-128, 128-256, 256-512, 512-1024},
	    xtick={1,2,3,4,5,6,7,8,9,10},
	    ylabel={\scriptsize Counting},
	    title={GraphSAGE},
	    ]
	\addplot[color=my_sin, fill=my_sin] coordinates
		{(1, 40) (2, 149) (3, 127) (4, 125) (5, 118) (6, 127) (7, 136) (8, 88) (9, 58) (10, 0)}
		\closedcycle;
	\end{axis}
\end{tikzpicture}
}
\resizebox{\linewidth}{!}{
\begin{tikzpicture}
	\begin{axis}[
		smooth,
		area style,
		enlarge x limits=false,
		xticklabels={0-2, 2-4, 4-8, 8-16, 16-32, 32-64, 64-128, 128-256, 256-512, 512-1024},
	    xtick={1,2,3,4,5,6,7,8,9,10},
	    ylabel={\scriptsize Counting},
	    title={CNE},
	    ]
	\addplot[color=flamingo, fill=flamingo] coordinates
		{(1,589) (2,173) (3,66) (4,36) (5,14) (6,11) (7,11) (8,30) (9,38) (10,0)}
		\closedcycle;
	\end{axis}
\end{tikzpicture}
}
  \caption{%
  The distribution of the similarity ranks of the subsequently clicked item.}
  \label{rank_distribution}
\end{figure}
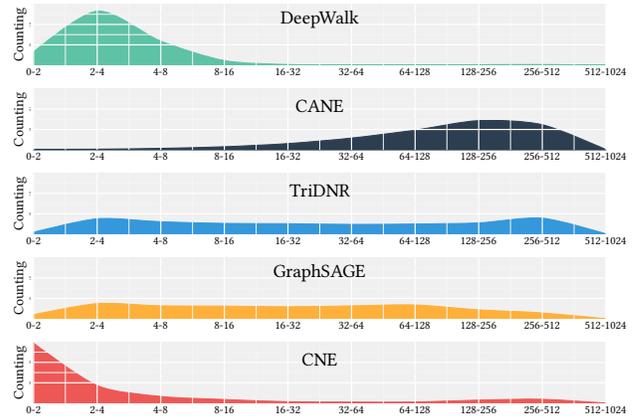

\textbf{Results } The distribution of the similarity ranks of the correct answers are presented in Fig.~\ref{rank_distribution}.
CNE tends to place the correct answer in the top three positions, and DeepWalk tends to place the correct answer in little bit farther positions.
Meanwhile, it can not be ignored that a bit of correct answers appear in much farther positions provided by CNE.
Besides, the other baselines that utilize node attributes also have the similar phenomenon.
The reason is that a lot of products with similar attributes but long geodesic distances may be ranked in the front of the list.
GraphSAGE, CANE, and TriDNR have very poor performances on this task.
The text feature seems to badly confuse their predictions.

\textbf{Case Study } We also check the items in the top positions calculated by different network embedding methods.
Table~\ref{casestudy} presents a case to illustrate their performance difference.
The first 4 lines are the product titles in the click records and the following items are recommended by different methods.
Note that the recommended product that has appeared in the records are filtered out.
The record indicates that the user aims at large size (\textit{loose}, \textit{plus size}), casual (\textit{casual}, \textit{floral printed}, \textit{flowy}), feminine (\textit{off shoulder}, \textit{princess}, \textit{pierced lace}) clothes in hot weather (\textit{T-shirt}, \textit{blouse}, \textit{dress}, \textit{3/4 sleeve}).
The top ranked items provided by the baselines only have partial features and most of them are identical words.
GraphSAGE and DeepWalk are not able to put similar products in the top positions.
CANE and TriDNR put the products with identical words in the top positions.
Compared with the record products, products recommended by CNE are with semantically similar titles although the words are different. 
The top ranked items have the corresponding features: large size (\textit{figure-flattering}, \textit{slimming}), casual (\text{flower}, \textit{colorful}, \textit{dotted}), feminine (\textit{princess}, \textit{fairy}, \textit{cold shoulder}, \textit{pink}), clothes in hot weather (\textit{dress}, \textit{silk}).
This case shows that CNE learns to capture the feature relevance and is capable of predict the users' interests.

\begin{table}
\caption{A user click record and the top ranked products provided by each method. Words of similar meaning are in the same color (\colorbox{size}{\makebox(36,4){large size}}, \colorbox{casual}{\makebox(24,4){casual}}, \colorbox{female}{\makebox(32,4){feminine}}, \colorbox{hot}{\makebox(44,4){hot weather}})\label{casestudy}}
\begin{adjustbox}{max width=0.9\linewidth}
\begin{tabular}{l|c|l}
\toprule
 & Rank & Product Title \\ \midrule
\multirow{4}{*}{Click Record}   & 1 & \colorbox{hot}{\makebox(24,4.5){Spring}} green \colorbox{size}{\makebox(20,4.5){loose}} \colorbox{hot}{\makebox(40,4.5){mid-sleeve}} casual T-shirt.       \\
                          & 2 & \colorbox{female}{\makebox(47,4.5){Pierced lace}} \colorbox{female}{\makebox(46,4.5){off shoulder}} \colorbox{hot}{\makebox(38,4.5){3/4 sleeve}} \colorbox{size}{\makebox(20,4.5){loose}} blouse.  \\
                          & 3 & \colorbox{size}{\makebox(34,4.5){Plus size}} \colorbox{casual}{\makebox(52,4.5){floral printed}} \colorbox{size}{\makebox(36,4.5){slimming}} \colorbox{female}{\makebox(32,4.5){princess}} dresses. \\
                          & 4 & Fake-two-piece \colorbox{female}{\makebox(46,4.5){pierced lace}} \colorbox{casual}{\makebox(20,4.5){flowy}} \colorbox{hot}{\makebox(15,4.5){tank}} blouse.      \\ \midrule
\multirow{3}{*}{DeepWalk} & 1 & Cotton plain \colorbox{size}{\makebox(18,4.5){loose}} white t-shirt.   \\
                          & 2 & \colorbox{hot}{\makebox(76,4.5){Spring and summer}} outlet \colorbox{female}{\makebox(42,4.5){high-waist}} shorts.       \\
                          & 3 & Ethnic style Thailand Napal \colorbox{hot}{\makebox(28,4.5){summer}} holiday long dress.  \\ \midrule
\multirow{3}{*}{CANE}     & 1 & Original design fashion \colorbox{size}{\makebox(18,4.5){loose}} hip pants. \\
                          & 2 & Ethnic style Thailand Napal \colorbox{hot}{\makebox(28,4.5){summer}} holiday long dress. \\
                          & 3 & \colorbox{hot}{\makebox(31,4.5){Summer}} \colorbox{hot}{\makebox(38,4.5){sleeveless}} wrinkled dress.  \\ \midrule
\multirow{3}{*}{TriDNR}   & 1 & Puff sleeve elegant \colorbox{casual}{\makebox(52,4.5){floral printed}} blouse.   \\
                          & 2 & \colorbox{size}{\makebox(38,4.5){Extra size}} \colorbox{size}{\makebox(34,4.5){slimming}} pierced long scarf wrap shawl.\\
                          & 3 & \colorbox{hot}{\makebox(76,4.5){Spring and summer}} \colorbox{hot}{\makebox(38,4.5){sleeveless}} \colorbox{casual}{\makebox(24,4.5){casual}} jumpsuits.   \\ \midrule
\multirow{3}{*}{GraphSAGE}& 1 & Korean \colorbox{hot}{\makebox(30,4.5){summer}} beautiful dress.  \\
						& 2 & Hong-kong \colorbox{female}{\makebox(40,4.5){embroidery}} dress.    \\
                        & 3 & Korean \colorbox{hot}{\makebox(30,4.5){summer}} fashion v-neck hoodie.    \\ 
                         \midrule
\multirow{3}{*}{CNE}      & 1 & \colorbox{hot}{\makebox(31,4.5){Summer}} \colorbox{casual}{\makebox(24,4.5){flower}} \colorbox{size}{\makebox(64,4.5){figure-flattering}} \colorbox{female}{\makebox(32,4.5){princess}} dress.     \\
                          & 2 & \colorbox{size}{\makebox(36,4.5){Slimming}} \colorbox{female}{\makebox(52,4.5){cold shoulder}} \colorbox{female}{\makebox(48,4.5){empire waist}} \colorbox{female}{\makebox(16,4.5){fairy}} dress.    \\
                          & 3 & \colorbox{female}{\makebox(16,4.5){Pink}} \colorbox{casual}{\makebox(32,4.5){colorful}} dotted \colorbox{hot}{\makebox(13,4.5){silk}} long-sleeve blouse.       \\ \bottomrule 
\end{tabular}
\end{adjustbox}
\end{table}

\section{Conclusion and Future Work}
In this paper, we open a new frontier in network embedding by introducing the ``\textit{principle of compositionally}'' to node embeddings.
To address the main limitations of the existing approaches, we proposed a novel approach that can efficiently generate embeddings for unseen nodes by combining their internal attributes.
Experiments on four subtasks demonstrate the effectiveness and generalization ability of compositional network embeddings.
 
In this work, we simply verify the feasibility of ``\textit{principle of compositionally}'' in networks with text attributes.
A number of extensions and potential improvements remain to be explored.
For instance, extending CNE to incorporate label information for downstream tasks; 
extending the encoders of CNE to model multiple attributes of nodes completely;
incorporating the attributes from neighbors to improve CNE.
A particularly interesting direction for future work is exploring the interpretability of CNE.

\bibliographystyle{ACM-Reference-Format}
\bibliography{CNE}

\end{document}